\documentclass[conference,a4paper]{IEEEtran}
\IEEEoverridecommandlockouts

\usepackage[hidelinks]{hyperref}
\usepackage[cmex10]{amsmath}%American Math Society(AMS) math formatting
\usepackage{amssymb,amsfonts}%AMS extra symbols and fonts
\usepackage{dblfloatfix}%fix double column figure ordering and placement

\usepackage[ruled,vlined]{algorithm2e}
\usepackage{graphicx}
\usepackage{graphics}
\usepackage{subcaption} 
\usepackage{multirow}
\graphicspath{{Figures/PDF/}{Figures/PNG/}}

\usepackage{booktabs}
\usepackage{siunitx}
\usepackage[numbers,compress]{natbib}
\usepackage{texnames}
\usepackage{bm,bbm}
\usepackage{orcidlink}
\usepackage{epsfig}
\usepackage{adjustbox}
\usepackage{setspace}

\begin{document}
% \doublespacing

\title{Multi-Task Learning with Multi-Annotation Triplet Loss for Improved Object Detection
}

\author{	\IEEEauthorblockN{Meilun Zhou\textsuperscript{1}\orcidlink{0000-0002-5891-3203}}
	\IEEEauthorblockA{\textit{University of Florida}\\
		Gainesville, FL, USA, 32611\\
		zhou.m@ufl.edu}
	\and
	\IEEEauthorblockN{Aditya Dutt\textsuperscript{2}\orcidlink{0000-0001-6602-4564}}
	\IEEEauthorblockA{\textit{University of Florida}\\
		Gainesville, FL, USA, 32611\\
		aditya.dutt@ufl.edu}
	\and
	\IEEEauthorblockN{Alina Zare\textsuperscript{1}\orcidlink{0000-0002-4847-7604}}
	\IEEEauthorblockA{\textit{University of Florida}\\
		Gainesville, FL, USA, 32611\\
		azare@ufl.edu}
    \and
    \IEEEauthorblockA{\textsuperscript{1}{Department of Electrical and Computer Engineering} \textsuperscript{2}{Department of Computer and Information Science and Engineering}}
% Department of Electrical and Computer Engineering \\ 
% Department of Computer and Information Science and Engineering
}

\maketitle
\begin{abstract}
Triplet loss traditionally relies only on class labels and does not use all available information in multi-task scenarios where multiple types of annotations are available. This paper introduces a Multi-Annotation Triplet Loss (MATL) framework that extends triplet loss by incorporating additional annotations, such as bounding box information, alongside class labels in the loss formulation. By using these complementary annotations, MATL improves multi-task learning for tasks requiring both classification and localization. Experiments on an aerial wildlife imagery dataset demonstrate that MATL outperforms conventional triplet loss in both classification and localization. These findings highlight the benefit of using all available annotations for triplet loss in multi-task learning frameworks.
\end{abstract}


\section{Introduction}
\label{sec:intro}

Multi-task learning is a learning paradigm where a single model is trained to simultaneously perform multiple tasks by using complementary task information from different task representations. Previously, multi-task learning frameworks have shown promise for optimizing classification and localization objectives \cite{wang2020boundary}. However, many existing loss functions like cross-entropy or mean squared error may lack the capacity to capture the nuanced interdependencies between these objectives \cite{hermans2017defense}. Triplet loss, a popular metric learning approach, excels at class separation in latent spaces and is highly effective for tasks requiring robust classification \cite{hoffer2015deep}. However, the emphasis and sole requirement of class similarity often limits the applicability of triplet loss in scenarios requiring precise localization. Triplet loss effectively separates classes but does not inherently incorporate any information derived from box annotations needed for accurate localization. \par

Recently, aerial imagery has emerged as a powerful tool for observing animals across vast and remote areas \cite{elmore2021evidence}. Wildlife detection is crucial for ecosystem management, biodiversity conservation, and environmental policy-making. Object detection algorithms that automatically detect objects are essential for analyzing large amounts of image data. This application provides a real-world problem for investigating adding different types of annotations to triplet loss because this task inherently involves two objectives: classifying detected objects into different categories and localizing them within images using bounding boxes. The complexity of natural habitats combined with variability in animal appearances and challenging conditions like occlusion or diverse lighting increases the difficulty of wildlife detection \cite{zhou2021improving}. Traditional loss functions, including cross-entropy and mean squared error, often fail to capture the nuanced interdependencies between classification and localization tasks \cite{hermans2017defense}. \par

This study introduces a Multi-Annotation Triplet Loss (MATL) designed to address these limitations. By integrating bounding box annotations alongside class labels, the proposed method constructs a latent space that balances both the demands of classification and localization. The addition of bounding box annotations specifically provides information about various features of the boxes, like size or symmetry. Evaluation of MATL on wildlife detection in aerial imagery highlights the approach’s ability to enhance the shared representation space for multi-task learning by effectively leveraging information from all available label sources.

\section{Proposed Methodology}
Traditional triplet loss approaches effectively group similar samples while separating dissimilar ones based solely on class labels. However, class labels may be insufficient for more complex tasks requiring both identification and localization. We propose a modified triplet loss function that incorporates the bounding box annotation information alongside class labels to create a latent space designed for the multi-task objectives of object classification and precise localization. The following subsections review the standard triplet loss formulation, detail our method of integrating bounding box annotations into our modified triplet loss, and outline the network architecture designed for wildlife detection in aerial imagery.

\subsection{Triplet Loss Formulation}
\label{sec:method}
The standard triplet loss is commonly used in metric learning to differentiate between embeddings of similar and dissimilar samples \cite{hoffer2015deep}. The loss operates on a set of three samples: an anchor sample \textit{a}, an positive sample \textit{p}, and a negative sample \textit{n}. For each anchor sample, the positives and negatives are determined using the class labels \textit{y}. Anchor and positive belong to the same class, while the negative belongs to a different class than the anchor. The triplet loss objective is to ensure that the distance between the anchor and positive embeddings is smaller than the distance between the anchor and negative embeddings by at least a margin $\alpha$. In this manner, triplet networks cluster different classes in the latent space by focusing on class information. \par

The loss function is formulated as follows:
\begin{equation}
\label{equation:basetriplet}
\begin{split}
  \mathcal{L}_{triplet}(a, p, n, y) = \max ( d(f(a), f(p))\\
  - d(f(a), f(n)) + \alpha, 0 ),
\end{split}
\end{equation}
\noindent where $f$ is the embedding function or feature extractor, $\alpha$ is the margin (set by the user) enforced between positive and negative samples, and $d$ is a distance function. 

\subsection{Multi-Annotation Triplet Loss Formulation}
We include bounding box characteristics as an additional annotation within the triplet loss framework to improve the model's ability to learn multi-task representations. While standard triplet loss uses only class labels to guide embedding separation, our approach integrates bounding box information by defining new discrete box labels based on the spatial characteristics of the target object's box annotations. Specifically, this study focuses on the addition of two key bounding box properties: box size and symmetric squareness. 

% \subsubsection{Extracting Box Features}
The box area feature was calculated by multiplying the width and height of each target object's bounding box, providing a direct measure of the object’s spatial extent. Symmetric squareness was determined by calculating the minimum of the width-to-height ratio and the height-to-width ratio, subtracted from $1$. A ratio closer to $0$ indicates a more symmetric or square-like box shape. This feature distinguishes between objects of different areas and symmetry to provide an additional dimension for separating the box features.

%% Edited figure. the figure above is original version
\begin{figure}[!htb]

\begin{minipage}{0.33\linewidth}
  \centering
  
\centerline{\epsfig{figure=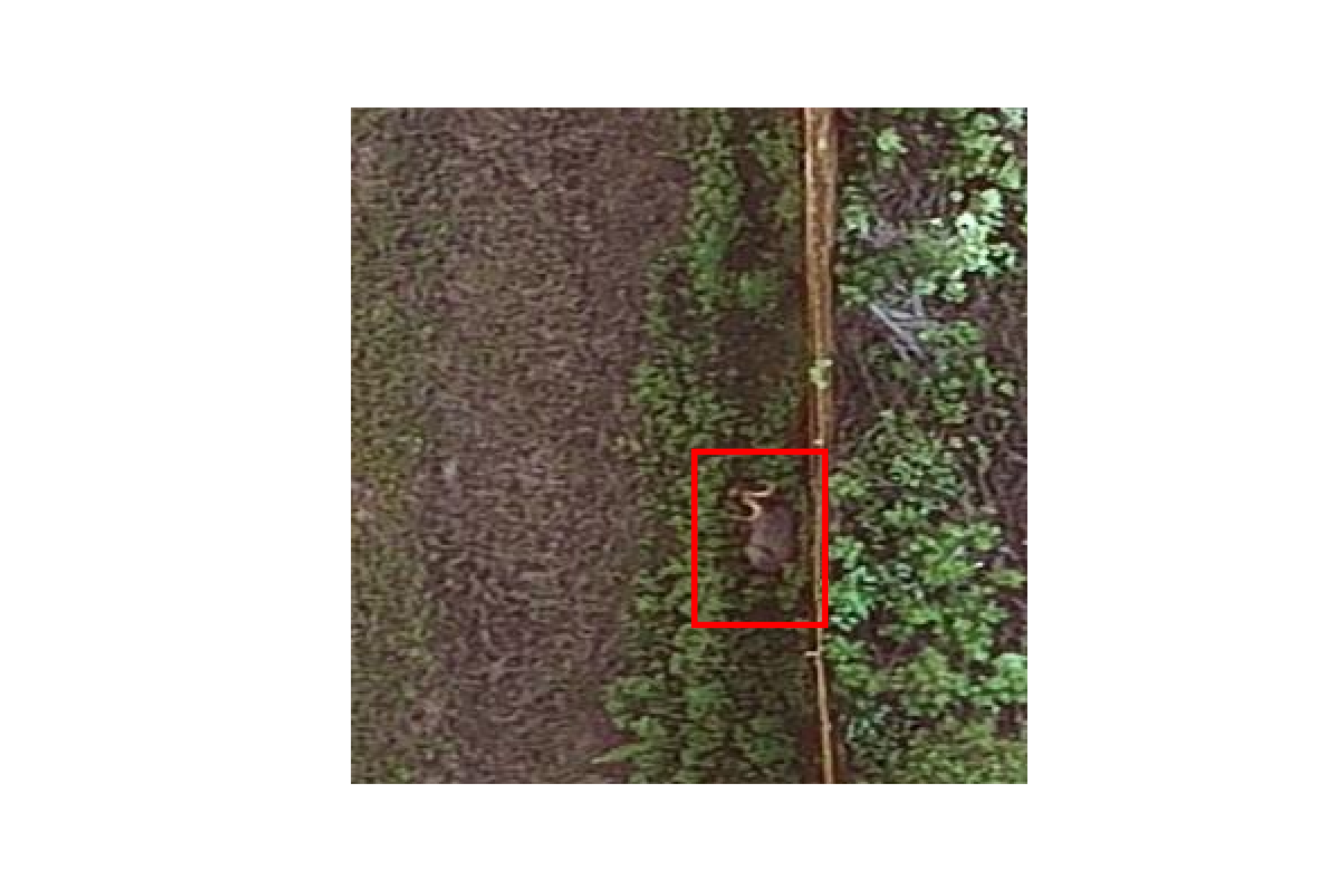},width=4cm}}
  % \vspace{0.02cm}
  \centerline{(a)}
\end{minipage}
\begin{minipage}{.33\linewidth}
  \centering
\centerline{\epsfig{figure=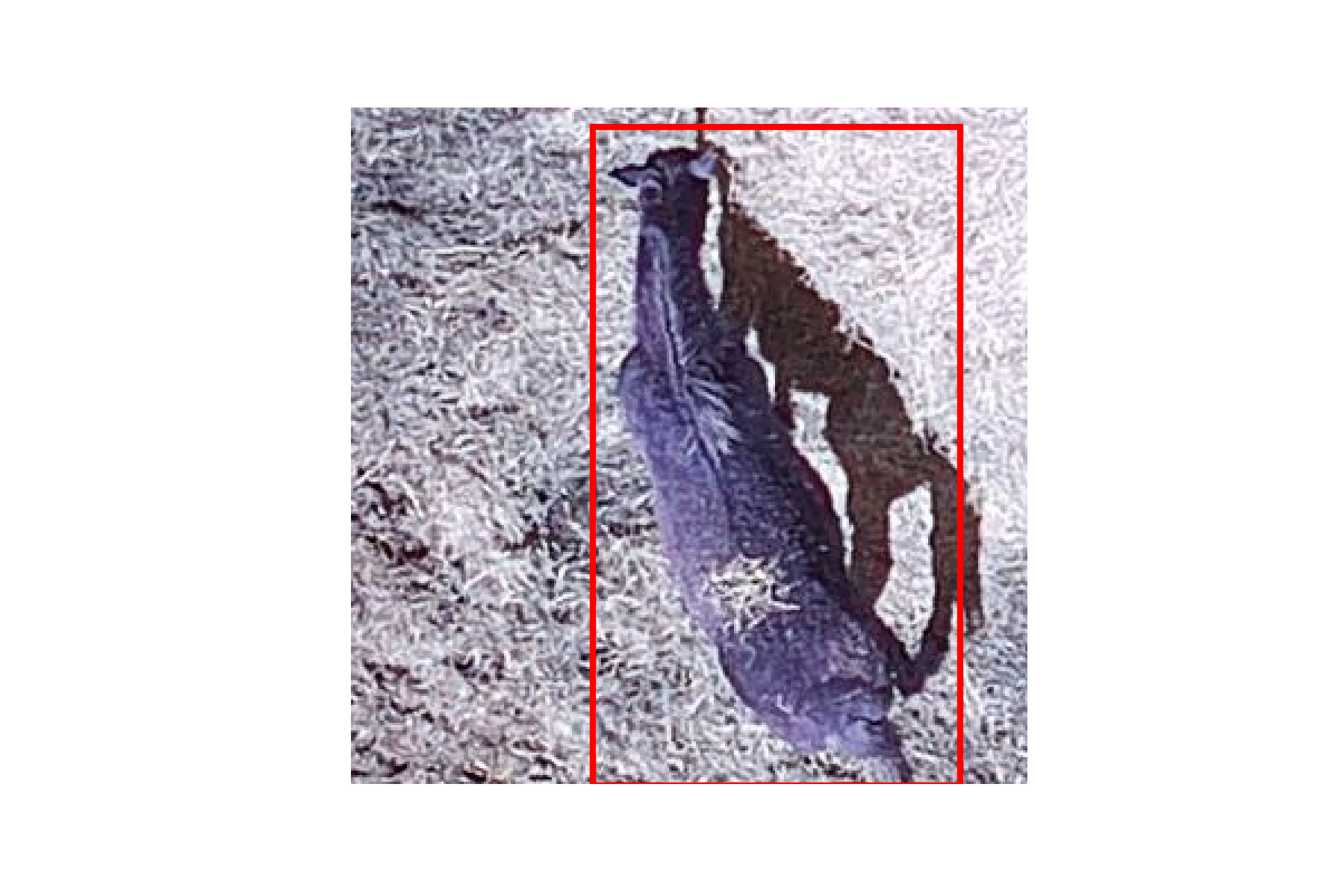,width=4cm}}
  % \vspace{0.02cm}
  \centerline{(b)}
\end{minipage}
\hfill
\begin{minipage}{0.29\linewidth}
  \centering
\centerline{\epsfig{figure=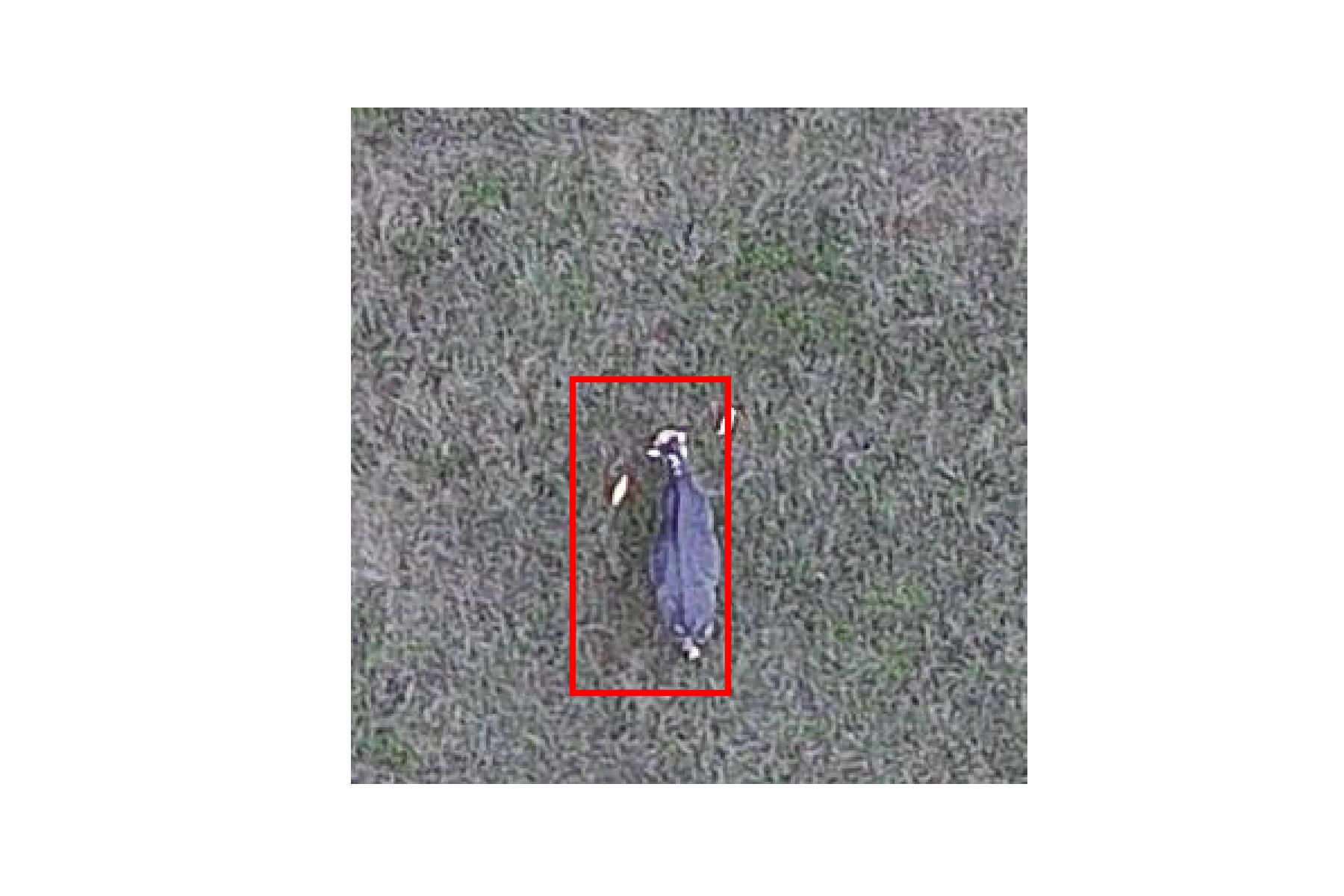,width=4cm}}
  % \vspace{0.02cm}
  \centerline{(c)}
  
\end{minipage}
\caption{Examples of three different box classes present in aerial imagery. (a) Deer, a small-square box, (b) Horse, large-elongated box, (c) Cow, small-elongated box.}
\label{fig:threebox}
\end{figure}

% \subsubsection{Clustering Box Information}
We then applied K-means clustering \cite{likas2003global} on the box area, symmetric squareness, width, and height values to categorize bounding box information. By clustering these features into three distinct groups, three new ``labels" that correspond to varying box size and squareness profiles were created. The new box labels, along with their corresponding RGB image, are shown in Fig. \ref{fig:threebox}. The number of clusters was chosen using the Elbow Method \cite{bholowalia2014ebk} where the within-cluster sum of squares displayed a distinct inflection point. These clusters capture common bounding box patterns across objects, enabling the model to generalize better to objects with similar spatial characteristics while maintaining distinctions in class. Prior to clustering, all features were independently normalized using Min-Max normalization to ensure uniform scaling and eliminate magnitude discrepancies across dimensions using the formula:  
\begin{equation}
\label{equation:minmaxnorm}
\begin{split}
    x' = \frac{x - \text{min}(x)}{\text{max}(x) - \text{min}(x)},
\end{split}
\end{equation}
\noindent where \(\text{min}(x)\) and \(\text{max}(x)\) represent the minimum and maximum values of the feature \(x\) across the dataset, respectively. This normalization step guarantees that all features contribute equally to the clustering objective, preventing dominance by features with larger ranges.

\begin{figure*}[!htb]
    \centering

    % Subfigure (a)
    \begin{subfigure}[b]{0.29\linewidth} % Adjust width as needed
        \centering
        \includegraphics[scale=0.38]{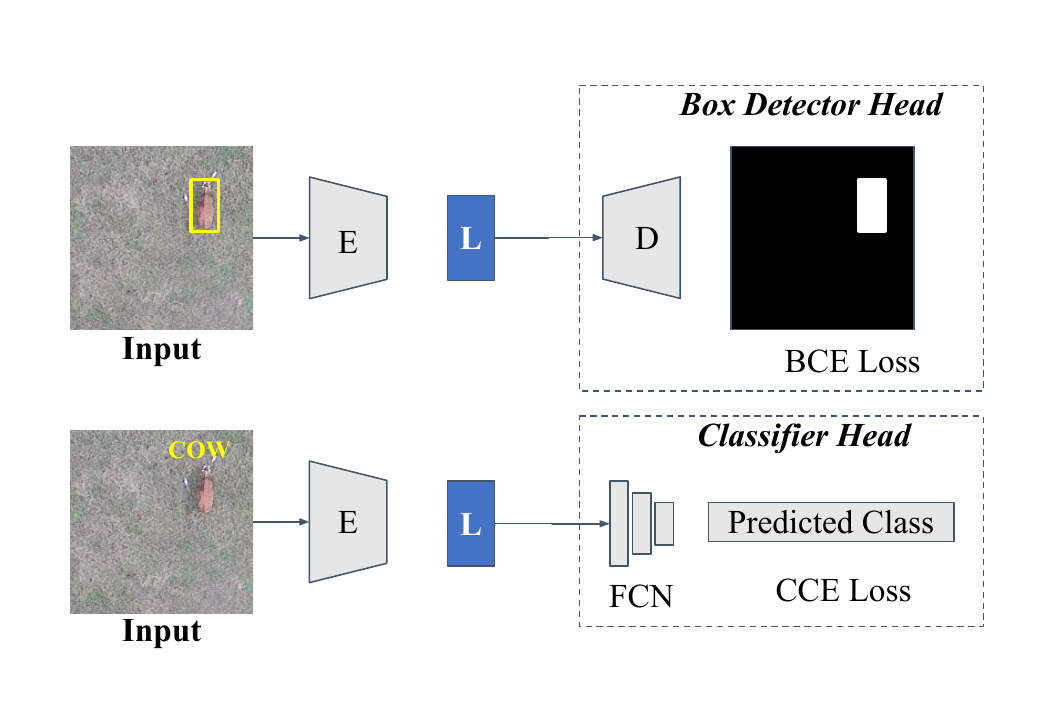}
        \caption{Single-Task model architecture.}
        \label{fig:awir_arch_a}
    \end{subfigure}
    \hspace{1.2cm}
    \begin{subfigure}[b]{0.63\linewidth} % Adjust width as needed
        \centering
        \includegraphics[scale=0.38]{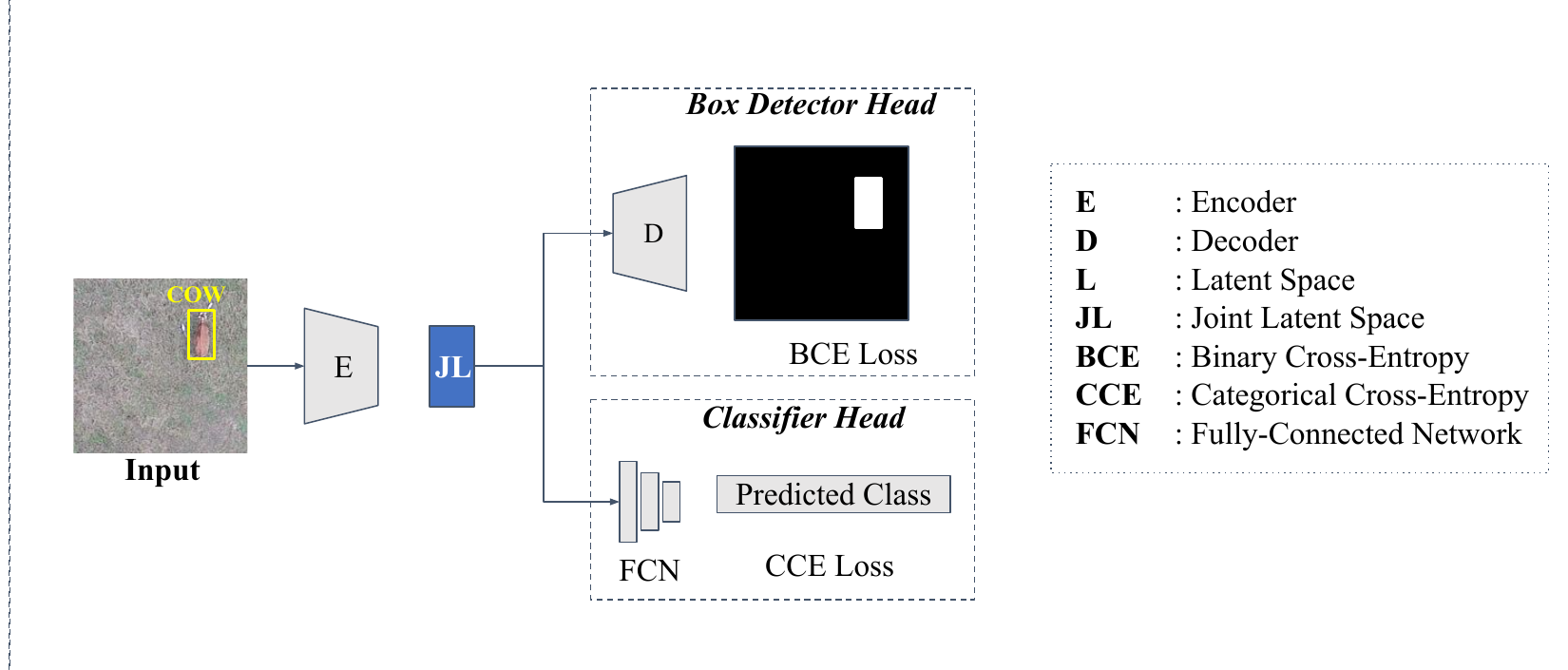}
        \caption{Multi-Task model architecture.}
        \label{fig:awir_arch_b}
    \end{subfigure}

    \caption{(a) Single-Task and (b)  Multi-Task model architectures.}
    \label{fig:awir_arch}
\end{figure*}

With the new box labels assigned to each sample, we first implemented a triplet loss focused on the box labels based off Equation \ref{equation:basetriplet}: 
\begin{equation}
\label{equation:boxtriplet}
\begin{split}
  \mathcal{L}_{box} = {L}_{triplet}(a, p, n, y_{box}),
\end{split}
\end{equation}
\noindent where $y_{box}$ are the three different types of box labels. 
Then, the typical class label triplet loss is formulated as: 
\begin{equation}
\label{equation:classtriplet}
\begin{split}
  \mathcal{L}_{class} = {L}_{triplet}(a, p, n, y_{class}),
\end{split}
\end{equation}
\noindent where $y_{class}$ are the class labels. Our proposed loss, Multi-Annotation Triplet Loss (MATL), combines the traditional class-based triplet loss with an additional triplet loss component that operates on these box labels. These two triplet losses shape the latent space simultaneously. The complete loss function is expressed as:
\begin{equation}
\label{equation:MATL}
\begin{split}
  \mathcal{L}_{MATL} = \left( 1 - \lambda \right) \mathcal{L}_{\text{class}} + \lambda  \mathcal{L}_{\text{box}},
\end{split}
\end{equation}

\noindent where the hyperparameter $\lambda$ controls the weight of the box label triplet loss relative to the class label loss. This hyperparameter allows for fine-tuning of the contribution from each label type. Setting $\lambda$ closer to $1$ emphasizes box label information, while lower values prioritize class differentiation. In this paper, we distinguish raw box annotations, the dataset's ground truth boxes, from user-generated box labels derived through K-means clustering.

\subsection{Network Architecture}
The network architecture in this study consists of an auto-encoder design aimed at extracting feature representations for two downstream tasks: classification and binary mask segmentation. The encoder extracts multi-scale features from images using dilated convolutions, starting with $16$ filters and increasing to $512$. The decoder reconstructs the binary mask by upsampling the encoded vector with transposed convolutions and residual connections. The final convolutional layers with sigmoid activation produce a mask, where pixel values over $0.5$ represent the target region. During inference, a bounding box is then fitted around the largest connected component. The classifier head is a fully connected network with four hidden layers. Batch normalization and ReLU activations stabilize training after each layer. The final layer uses softmax activation to output probabilities for three target classes. 

In the multi-task setup, the embeddings produced by the encoder serve as a shared representation for multiple task-specific heads in parallel. Each task-specific head processes these joint embeddings independently to predict outcomes for its respective task. This design allows the encoder to learn features that generalize across all tasks in order to leverage complementary information from multiple objectives. In contrast, the single-task setup restricts the encoder embeddings to flow into a single task-specific head, which focuses the feature extraction entirely on the needs of the individual task. While single-task setups may optimize performance for the single task, the restriction may limit the encoder’s ability to learn features that capture broader, cross-task relationships. The setups are displayed in Fig. \ref{fig:awir_arch}. \par

All models were trained with Tensorflow on an NVIDIA A100 GPU with $12$ CPUs and $60$ GB of memory allocated. This configuration provided the computational resources necessary for efficient training and optimization. Code can be found at \url{https://github.com/GatorSense/MATL/tree/main}.

\section{Experimental Results}
\label{results}

The Animal Wildlife Image Repository (AWIR) \cite{samiappan2024aerial} is a dataset created to support research in wildlife detection, classification, and monitoring. AWIR contains a collection of RGB images capturing various species of animals in their natural environments. We first cropped images containing animals into smaller tiles of size $300\times300$ pixels. Each tile includes only a single animal, which simplifies the model's task by focusing attention on a single target per image. To avoid bias in the object’s location within each tile, we employed a random cropping strategy around the animal that ensures that the animal is not always centered. Therefore, the model learns to detect the object regardless of position in the image. All images were normalized using the same MinMax normalization in Equation \ref{equation:minmaxnorm} before passing into the network. The box annotations were clustered and the mean values of the area and symmetric squareness features from each cluster are further detailed in Table \ref{table:boxfeatures}. \par

\begin{table}[!htb]
\caption{The mean normalized area and symmetric squareness (SS) values used in bounding box portion of the Multi-Annotation Triplet Loss.}
\begin{center}
\begin{tabular}{||c | c | c||} 

 \hline
 \textbf{Category} & \textbf{Norm. Area} & \textbf{Norm. SS} \\ [0.5ex] 
 \hline\hline
\textbf{Small-Elongated Box} & 0.088 & 0.621 \\ 
 \hline
\textbf{Large-Elongated Box} & 0.630  & 0.559  \\
 \hline
\textbf{Small-Square Box} & 0.093 & 0.173  \\
 \hline

% \hline\hline
% Small Elongated Box & 0.088 & 0.070 & 0.621 & 0.139 \\ 
%  \hline
% Large Elongated Box & 0.630 & 0.127 & 0.559 & 0.203 \\
%  \hline
% Small Square Box & 0.093 & 0.065 & 0.173 & 0.111 \\
%  \hline
 
\end{tabular}

\end{center}
\label{table:boxfeatures}
\end{table}
\begin{table*}[htb]
\caption{Results of Single-Task (ST) and Multi-Task (MT) models on classification and box localization tasks are presented. Multi-Annotation Triplet Loss (MATL) is compared with Without Triplet Loss (WTL) and Class Label Triplet Loss (CLTL) across both models and tasks. MATL results with box triplet loss weights ($\lambda$) of $0.25$, $0.5$, and $0.75$ are reported. For classification, mean overall accuracy (\%) and standard deviation across eight experiments are reported; for box localization, mean IoU and standard deviation are provided.}

% \begin{center}
% \begin{tabular}{||c | c | c | c | c | c | c||} 
%  \hline
%  \multirow{2}{*}{\textbf{Model}}  & \multicolumn{3}{c}{\textbf{Classification (\textit{Overall Accuracy)}}} & \multicolumn{3}{|c||}{\textbf{Box Localization \textit{(IoU)}}} \\ [0.5ex] 
%  \cline{2-7}
%   & \textbf{WTL} & \textbf{CLTL} & \textbf{MATL} \textbf{\textit{(0.25)}}  & \textbf{WTL} & \textbf{CLTL} & \textbf{MATL} \textbf{\textit{(0.25)}} \\ [0.5ex] 

%  \hline\hline
%  \textbf{Single-Task} & 0.58 $\pm$ 0.06  & 0.73 $\pm$ 0.01 & 0.83 $\pm$ 0.02  & 0.19 $\pm$ 0.03 & 0.16 $\pm$ 0.02  & 0.18 $\pm$ 0.02  \\ 
%  \hline
%  \textbf{Multi-Task} & 0.68 $\pm$ 0.04 & 0.78 $\pm$ 0.05 & 0.83 $\pm$ 0.02 &  0.18 $\pm$ 0.01 & 0.15 $\pm$ 0.03  & 0.19 $\pm$ 0.02  \\
%  \hline
%  % ST/LOC & 0.19 $\pm$ 0.03 & 0.16 $\pm$ 0.02  & 0.18 $\pm$ 0.02  \\
%  % \hline
%  % MT/LOC & 0.18 $\pm$ 0.01 & 0.15 $\pm$ 0.03  & 0.19 $\pm$ 0.02  \\
%  % \hline
% \end{tabular}
% \end{center}
% \label{table:results}
% \end{table*}

\setlength{\tabcolsep}{3pt} 
\begin{center}
% \small
\adjustbox{max width=\textwidth}{ 
\begin{tabular}{||c | c | c | c | c | c | c | c | c | c | c||}
 \hline
 \multirow{3}{*}{\textbf{Model}}  & \multicolumn{5}{c}{\textbf{Classification (Accuracy)}} & \multicolumn{5}{|c||}{\textbf{Box Localization (Intersection over Union)}} \\ [0.5ex] 
 \cline{2-11}
  & \multirow{2}{*}{\textbf{WTL}}  & \multirow{2}{*}{\textbf{CLTL}}  & \multirow{2}{*}{\textbf{MATL}}   & \multirow{2}{*}{\textbf{MATL}} & \multirow{2}{*}{\textbf{MATL}}   & \multirow{2}{*}{\textbf{WTL}}  & \multirow{2}{*}{\textbf{CLTL}}  & \multirow{2}{*}{\textbf{MATL}} & \multirow{2}{*}{\textbf{MATL}}  & \multirow{2}{*}{\textbf{MATL}}  \\ [0.5ex] 
  &  &  & \textbf{\textit{(0.25)}} &  \textbf{\textit{(0.5)}} & \textbf{\textit{(0.75)}}  &  &  &  \textbf{\textit{(0.25)}} &  \textbf{\textit{(0.5)}} &  \textbf{\textit{(0.75)}} \\ [0.5ex] 

 \hline\hline
 \textbf{Single-Task} & $58.2 \pm 6.1$  & $73.3 \pm 1.5$ & $\mathbf{83.0 \pm 2.3}$  & $79.4 \pm 2.6$ & $79.9 \pm 4.3$  & $\mathbf{0.190 \pm 0.025}$ & $0.157 \pm 0.022$ & $0.184 \pm 0.017$ & $0.185 \pm 0.018$ & $0.185 \pm 0.015$  \\ 
 \hline
 \textbf{Multi-Task} & $67.8 \pm 3.9$ & $78.1 \pm 5.0$ & $\mathbf{83.2 \pm 1.7}$ &  $81.9 \pm 2.6$ & $80.1 \pm 3.0$  & $0.178 \pm 0.013$ & $0.149 \pm 0.025$ & $\mathbf{0.189 \pm 0.016}$ & $0.177 \pm 0.018$ & $0.179 \pm 0.012$  \\
\hline

 %  \textbf{ST} & $\left[57.3, 59.1\right]$  & $\left[73.1, 73.5\right]$ & $\mathbf{\left[82.7, 83.4\right]}$  & $\left[79.0, 79.8\right]$ & $\left[79.3, 80.6\right]$  & $\left[0.186, 0.194\right]$ & $\left[0.154, 0.160\right]$ & $\left[0.181, 0.187\right]$ & $\mathbf{\left[0.182, 0.188\right]}$ & $\left[0.183, 0.187\right]$  \\ 
 % \hline
 % \textbf{MT} & $\left[67.2, 68.4\right]$ & $\left[77.3, 78.9\right]$ & $\mathbf{\left[82.9, 83.5\right]}$ &  $\left[81.5, 82.3\right]$ & $\left[79.7, 80.6\right]$  & $\left[0.176, 0.180\right]$ & $\left[0.145, 0.153\right]$ & $\mathbf{\left[0.187, 0.191\right]}$ & $\left[0.174, 0.180\right]$ & $\left[0.177, 0.181\right]$  \\
 % \hline

 % ST/LOC & 0.19 $\pm$ 0.03 & 0.16 $\pm$ 0.02  & 0.18 $\pm$ 0.02  \\
 % \hline
 % MT/LOC & 0.18 $\pm$ 0.01 & 0.15 $\pm$ 0.03  & 0.19 $\pm$ 0.02  \\
 % \hline
\end{tabular}
}
\end{center}
\label{table:results}
\end{table*}

We compared the multi-task model’s performance to the models without the triplet loss and with the standard class-label-based triplet loss, while keeping all model architectures identical. The classification and box localization tasks are evaluated using the overall classification accuracy and Intersection over Union (IoU), respectively. All models were trained and validated eight different times with $30$\% of the total data using a stratified K-fold technique where $K = 8$. These models were then tested on the remaining $70$\% of the data. Average results of the experiments are shown in Table \ref{table:results}.

% \begin{table}[!htb]

% \caption{Results from Single-Task (ST) and Multi-Task (MT) classification (Cls.) and box localization (Loc.) models on testing set with three different embedding spaces: Without Triplet Loss (WTL), Class Label Triplet Loss (CLTL), and Multi-Annotation Triplet Loss (MATL) at 0.25 weight for the box label triplet. Classification results are reported as classification accuracy along with standard deviation over 8 runs. Localization results are reported as IoU along with standard deviation over 8 runs.}
% \begin{center}
% \begin{tabular}{||c | c | c | c||} 
%  \hline
%  Experiment & WTL & CLTL & MATL(0.25) \\ [0.5ex] 
%  \hline\hline
%  ST Cls. & 0.58 $\pm$ 0.06  & 0.73 $\pm$ 0.01 & 0.83 $\pm$ 0.02  \\ 
%  \hline
%  MT Cls. & 0.68 $\pm$ 0.04 & 0.78 $\pm$ 0.05 & 0.83 $\pm$ 0.02  \\
%  \hline
%  ST Loc. & 0.19 $\pm$ 0.03 & 0.16 $\pm$ 0.02  & 0.18 $\pm$ 0.02  \\
%  \hline
%  MT Loc. & 0.18 $\pm$ 0.01 & 0.15 $\pm$ 0.03  & 0.19 $\pm$ 0.02  \\
%  \hline
% \end{tabular}
% \end{center}
% \label{table:results}
% \end{table}

\begin{figure}[!htb]

    \centering
    \includegraphics[scale=0.62]{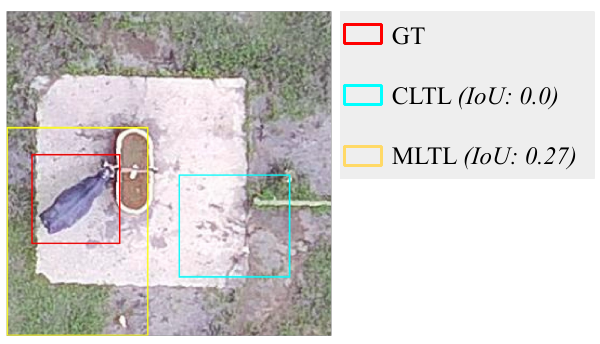}
    \caption{An example of the improved box localization from MATL as compared to CLTL for class cow.}
    \label{fig:CLTLvsMATL}

    % \hspace{1.2cm}

\end{figure}

We empirically determined the best weight $\lambda$ for MATL for this dataset to be $0.25$ after testing values of $0.25$, $0.5$, and $0.75$. For the classification task, the MATL consistently outperforms both models without triplet loss and the class-label triplet loss in single and multi-task setups. For the box detection task, the MATL at $0.25$ weight performs comparably to without triplet loss and demonstrates clear improvement over the class-label triplet loss. An example of this improvement in localization is shown in Fig. \ref{fig:CLTLvsMATL}. 

\section{Discussion}
The new latent space supervised by both class and box labels proves beneficial for multi-task, since both the classification and box tasks are improved, whereas the classification label embedding space shows a classification task dominance. By using both class label and box annotation supervision, the MATL is able to better leverage information from both tasks and outperform the models without triplet loss and class label triplet loss. The effect of the weight $\lambda$ for the box annotations on the overall MATL is important to consider since classification performance decreases as $\lambda$ increases. The result is an improvement over traditional single-task methods \cite{lakkapragada2023mitigating} and class-label triplet loss, which tend to focus on one task at the expense of the other.

\begin{figure}[htb]

\begin{minipage}[b]{0.3\linewidth}
  \centering
\centerline{\epsfig{figure=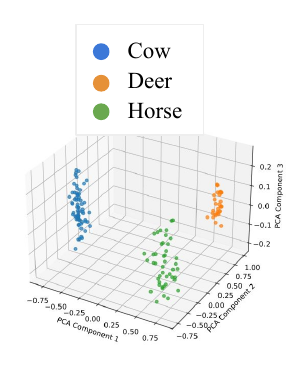,width=3.0cm}}
  \vspace{0.02cm}
  \centerline{(a)}\medskip
\end{minipage}
\quad
\begin{minipage}[b]{.3\linewidth}
  \centering
\centerline{\epsfig{figure=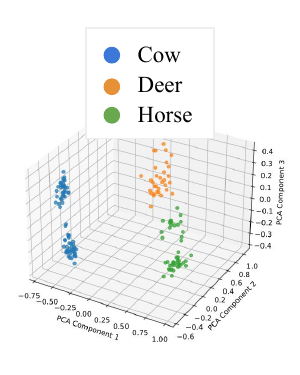,width=3.0cm}}
  \vspace{0.02cm}
  \centerline{(b)}\medskip
\end{minipage}
\hfill
\begin{minipage}[b]{0.3\linewidth}
  \centering
\centerline{\epsfig{figure=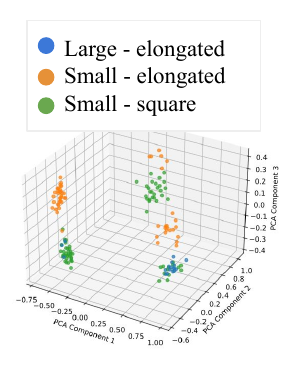,width=3.0cm}}
  \vspace{0.02cm}
  \centerline{(c)}\medskip
\end{minipage}

\caption{A comparison between different representations after applying Principal Component Analysis (PCA) is shown. (a) Class-label embeddings colored by class labels, (b) Multi-Annotation embeddings colored by class labels, and (c) Multi-Annotation embeddings colored by box labels.}
\label{fig:embspace}
\end{figure}

% \textcolor{red}{Add some discussion about the figure and how the clusters are formed. Add caption to the figure}

Fig. \ref{fig:embspace} (a) provides a comparative visualization of the latent space representations learned under different configurations of the triplet loss. In Fig. \ref{fig:embspace} (a), where the triplet loss is based solely on class labels, the embeddings form distinct clusters, with each cluster corresponding to a specific class. This demonstrates that the learned representation effectively captures inter-class separability. However, this approach assumes uniformity among samples within each class, neglecting any inherent intra-class variations that may be essential for certain tasks like object localization. In contrast, Fig. \ref{fig:embspace} (b) illustrates the latent space when both class and bounding box labels are incorporated into the triplet loss. The embeddings are colored by the class labels to show the variations within each class. The resulting latent space exhibits sub-clusters within each class cluster, reflecting variations among samples of the same class. These sub-clusters capture meaningful differences, such as variations in bounding box dimensions (\textit{e.g.}, width and height), which are critical for accurate localization. This suggests that samples within the same class are not always identical in their spatial properties, and these differences are naturally embedded in the latent representation. A similar pattern can be observed in Fig. \ref{fig:embspace} (c) when the embeddings are colored by box labels: ``large-elongated box", ``small-elongated box," and ``small square box." The ``small-elongated box" class forms multiple clusters within all class clusters because this type of box is present in all three animal classes. Similarly, ``small-square box" is present mostly in deer and cow classes. The horse class consists mostly of ``large-elongated box" which is indicated by the less scattered single cluster. \par

To summarize, by leveraging multiple attributes for each sample, the model captures a more comprehensive representation of the data. This approach maps similar samples closer together based on several characteristics, rather than relying on a single label. The resulting latent space is more robust and improves performance across tasks by considering the diverse aspects of each sample.

\section{Conclusion}
\label{sec:conclusion}

The proposed Multi-Annotation Triplet Loss improves multi-task scenarios by encouraging a more structured and meaningful representation of the data. This approach enhances the model's ability to handle multiple tasks simultaneously by encouraging the model to learn features that are shared across different types of labels. One direction moving forward is to explore a multi-modality approach \cite{dutt2022shared}, incorporating thermal data alongside other data types. Thermal data can provide complementary information, especially in low-light conditions, so combining thermal imagery with high-resolution RGB imagery could improve the model's robustness and accuracy \cite{krishnan2023fusion}. Additionally, instead of forcing both labels to share a single latent space, future work will investigate methods of combining information from two different labels, such as using concatenation or other effective fusion strategies. This would allow for more flexible and task-specific representations.

\small
\bibliographystyle{IEEEtranN}
\bibliography{references}

\end{document}